\documentclass{article}
\usepackage{spconf,amsmath,graphicx}

\usepackage{url}
\usepackage{hyperref}
\usepackage{makecell}
\usepackage{booktabs}
\usepackage{xcolor}  
\usepackage{colortbl}  
\usepackage{bbding}
\usepackage{multirow}

\title{Joint Multi-Person Body Detection and Orientation Estimation via One Unified Embedding}  

\name{Huayi Zhou$^{\star}$ \qquad Fei Jiang$^{\dagger}$ \qquad Jiaxin Si$^{\star}$ \qquad Hongtao Lu$^{\star}$}
\address{$^{\star}$ Shanghai Jiao Tong University, sjtu\_zhy@sjtu.edu.cn, sijiaxin@sjtu.edu.cn, htlu@sjtu.edu.cn \\
    $^{\dagger}$ East China Normal University, fjiang@mail.ecnu.edu.cn }

\begin{document}
%
\maketitle
%

\begin{abstract}
Human body orientation estimation (HBOE) is widely applied into various applications, including robotics, surveillance, pedestrian analysis and autonomous driving. Although many approaches have been addressing the HBOE problem from specific under-controlled scenes to challenging in-the-wild environments, they assume human instances are already detected and take a well cropped sub-image as the input. This setting is less efficient and prone to errors in real application, such as crowds of people. In the paper, we propose a single-stage end-to-end trainable framework for tackling the HBOE problem with multi-persons. By integrating the prediction of bounding boxes and direction angles in one embedding, our method can jointly estimate the location and orientation of all bodies in one image directly. Our key idea is to integrate the HBOE task into the multi-scale anchor channel predictions of persons for concurrently benefiting from engaged intermediate features. Therefore, our approach can naturally adapt to difficult instances involving low resolution and occlusion as in object detection. We validated the efficiency and effectiveness of our method in the recently presented benchmark MEBOW with extensive experiments. Besides, we completed ambiguous instances ignored by the MEBOW dataset, and provided corresponding weak body-orientation labels to keep the integrity and consistency of it for supporting studies toward multi-persons. Our work is available at \url{https://github.com/hnuzhy/JointBDOE}.
\end{abstract}


\section{Introduction}\label{intro}

The human body orientation estimation (HBOE) task is defined as estimating the skeleton orientation of one person at the orthogonal camera frontal view. It can not only be directly applied in many industrial applications, e.g., pedestrian behavior analysis in intelligent vehicles \cite{ rehder2014head, yu2019continuous} and attention estimation in classrooms \cite{araya2021automatic}, but also serve as a vital auxiliary for assisting other closely related upstream vision tasks \cite{ricci2015uncovering, raza2018appearance, wu2020mebow}. As a standalone problem, HBOE has been studied for a lone time \cite{andriluka2010monocular, baltieri2012people, choi2016human, hara2017growing, liu2017weighted, yu2019continuous}. Earlier, the widely used HBOE dataset TUD \cite{andriluka2010monocular} is built with eight coarse-grained orientation classes. Then, Hara et.al. \cite{hara2017growing} refined orientation labels of the TUD dataset into continuous angles. Using RGB-D sensors, the MCG-RGBD dataset \cite{liu2013accurate} can provide RGB images and depth information for achieving fine-grained HBOE.  Recently, MEBOW \cite{wu2020mebow} created a large-scale, high-precision, diverse-background dataset based on COCO \cite{lin2014microsoft} with its readily available human instances bounding box labels. This new benchmark with contextual information and the variety of background poses many unresolved real-world challenges for in-the-wild HBOE task.

\begin{figure}[]
	\centerline{\includegraphics[width=\columnwidth]{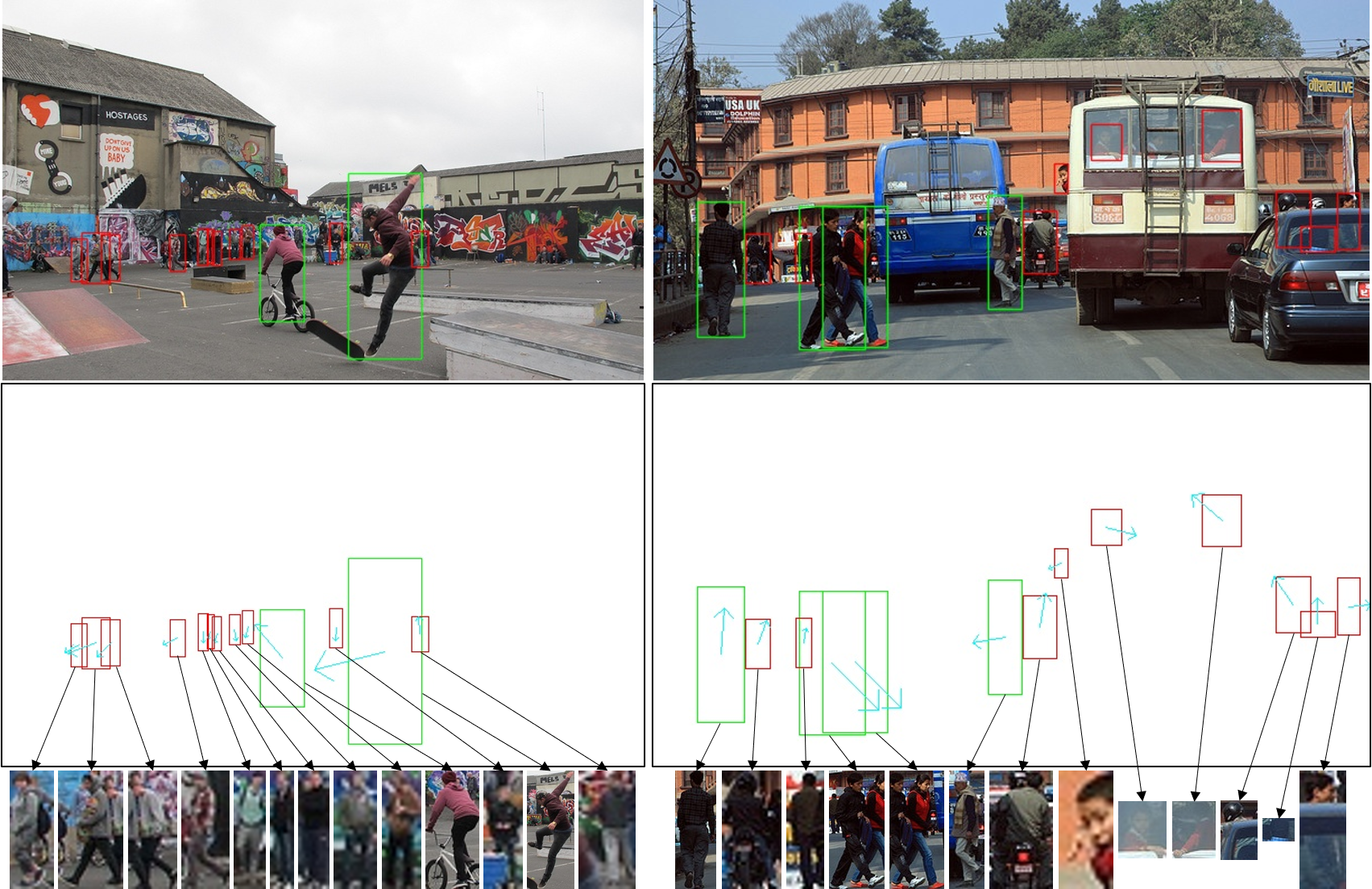}}
	\caption{Two examples of reconstructed MEBOW dataset \cite{wu2020mebow}. {\bf Top}: Original images with crowds. Green boxes are provided by MEBOW. Red boxes are newly added by us. {\bf Middle}: Corresponding body bounding box and orientation (in cyan arrow) labels.  {\bf Bottom}: Cropped human instances for verification. They may be truncated, occluded or tiny bodies, which are challenging for single HBOE methods yet reasonable for our single-stage HBOE approach regarding multi-persons.}
	\label{examples}
\end{figure}

However, most previous work in HBOE assumes the input is a well cropped human instance. When in real application, these single HBOE methods have to firstly obtain human instances by the pre-trained person detector (e.g., Faster R-CNN \cite{ren2015faster}). We argue this setting is defective facing multi-persons for two reasons. First, two separated stages will take linearly growing time as the number of people in the image increases. Second, one cropped instance may be incomplete or have other persons which will greatly interfere with feature discrimination. These two drawbacks also have been commonly referred and addressed in single-stage object detection \cite{tian2019fcos} and bottom-up human pose estimation \cite{cao2017realtime}.

To alleviate the shortcomings of two-stage multi-person HBOE methods, we propose a single-stage YOLOv5 \cite{jocher2020yolov5} based method for tackling the multi-person HBOE problem. Inspired by multi-task learning frameworks \cite{zhang2016joint, wu2021synergy, burgermeister2022pedrecnet, nonaka2022dynamic}, our method can jointly detect person and estimate body orientation. Specifically, we design a unified embedding including information of both bounding box and direction angle suitable for each anchor channel prediction, and selectively optimize body orientation using proposals with high probability of person instances. Benefiting from the single-stage object detection network, our method has the potential to be much faster than its two-stage counterparts, and performs better in crowded scenes. It is most probably preferred in real application for its simplicity and efficiency. In experiments, to adapt training of multi-persons, we reconstructed the annotation of MEBOW by introducing full body bounding boxes and weak orientation labels. Some examples are shown in Figure \ref{examples}.

Our contributions are three-folds: (1) We propose a novel single-stage framework for dealing with the multi-person HBOE task for the first time. It can realize person detection and body orientation estimation jointly. (2) We design a new generic unified embedding which extends the traditional object representation and associates the orientation angle into it. (3) Our method has achieved superior performance in reconstructed multi-person HBOE dataset MEBOW, which demonstrates its potential value in real application.

\section{Related Work}

\subsection{Human Body Orientation Estimation}
As a precondition step of human body orientation estimation (HBOE), human body detection is usually accomplished as a by-product using general object detectors trained on large-scale benchmarks like COCO \cite{lin2014microsoft}, or resolved by dedicated person detectors trained on pedestrian datasets (e.g., TUD \cite{andriluka2010monocular} and CityPersons \cite{zhang2017citypersons}) and human-oriented datasets (e.g., CrowdHuman \cite{shao2018crowdhuman}). The detection foundation has also progressed from the traditional hand-crafted feature-based approaches to the current deep learning-based methods including Faster R-CNN \cite{ren2015faster}, FCOS \cite{tian2019fcos} and YOLOv5 \cite{jocher2020yolov5}.

Similarly, previous studies \cite{andriluka2010monocular, baltieri2012people, liu2013accurate, hara2017growing} for HBOE tend to prefer feature engineering and traditional classifiers when the datasets having limited scales and coarse orientation labels. Methods \cite{choi2016human, raza2018appearance} tentatively applied simple multi-layer neural networks to solve the HBOE as a classification problem. Hara et.al. \cite{hara2017growing} re-annotated the TUD dataset with continuous orientation labels for predicting fine-grained orientations. Recently, MEBOW \cite{wu2020mebow} pushed deep learning-based HBOE methods forward by presenting a new large-scale challenging benchmark, and established a strong baseline model for HBOE. PedRecNet \cite{burgermeister2022pedrecnet} was a new multi-task network that supports various pedestrian-oriented functions including the HBOE which obtained a comparable result to the state-of-the-art \cite{wu2020mebow}. However, none of these methods has tried to solve the HBOE problem under multi-person scenarios. In this paper, to the best of our knowledge, we are the first to explore this problem with uncropped original frame as the input.


\subsection{Multi-task Learning}

Generally, multi-task learning strategies \cite{zhang2016joint, wu2021synergy} are favored for their high efficiency and potential of exploiting synergies among correlated tasks. The HBOE task has also been associated with estimations of head pose \cite{raza2018appearance, nonaka2022dynamic}, 3D human pose \cite{wu2020mebow, burgermeister2022pedrecnet} and 3D eye gaze \cite{araya2021automatic, nonaka2022dynamic}. Raza et.al. \cite{raza2018appearance} designed two parallel trained CNN classifiers for pedestrian head-pose and body-orientation. MEBOW \cite{wu2020mebow} treated body orientation as a lower-cost supervision source for assisting better 3D human pose estimation. PedRecNet \cite{burgermeister2022pedrecnet} proposed a unified multi-task architecture for full 3D human pose and orientation estimation. GAFA \cite{nonaka2022dynamic} introduced a new 3D gaze estimation method and dataset leveraging the intrinsic gaze, head, and body coordination of people. Again, the input of all these multi-task studies is an RGB cropped bounding box image of a human. Different from them, the approach we proposed for joint body detection and orientation estimation will remedy Section \ref{intro} mentioned inherent deficiencies of existing HBOE methods that take cropped single person as the input.


\begin{figure}[]
	\centerline{\includegraphics[width=\columnwidth]{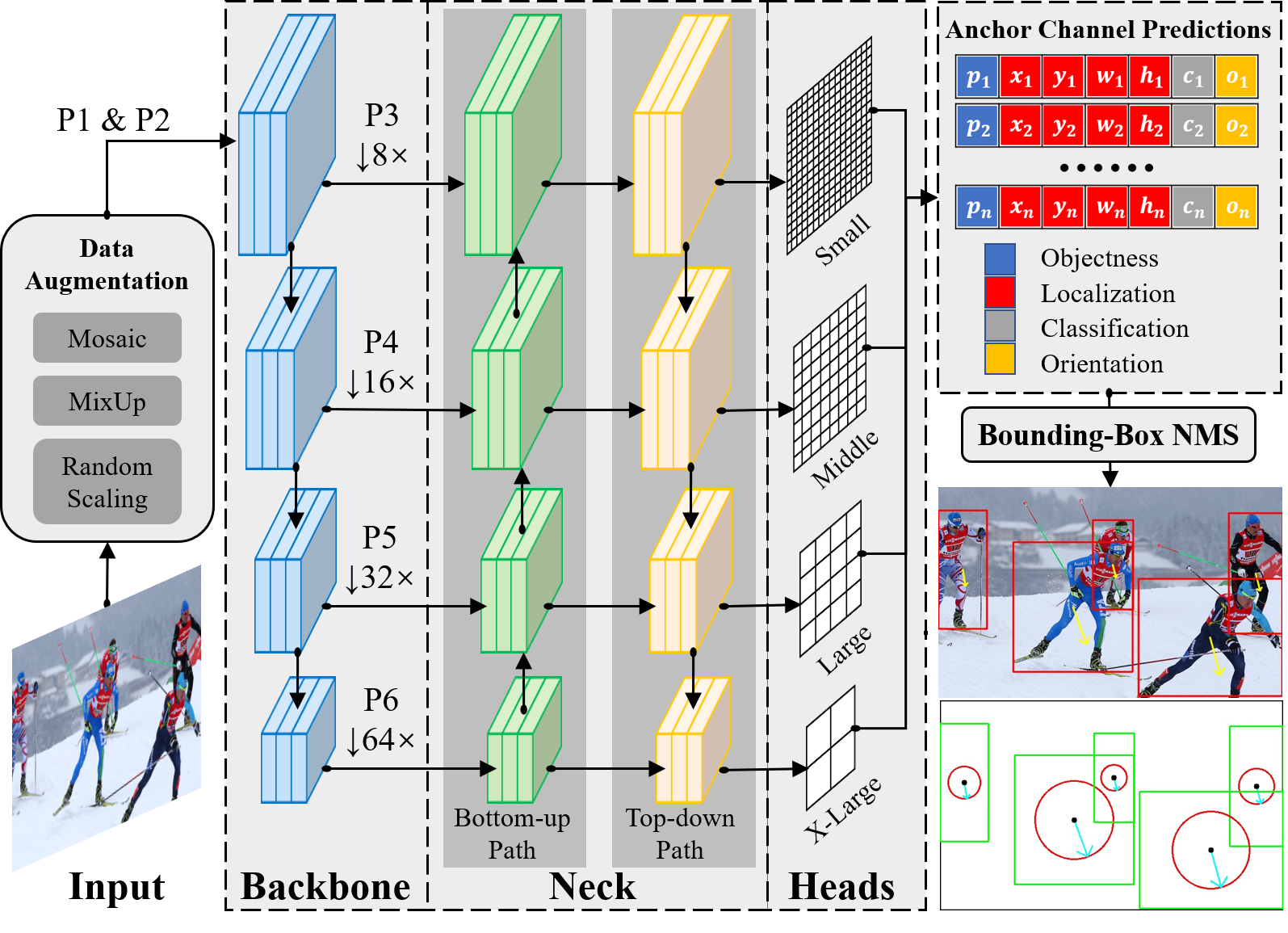}}
	\caption{Our architecture is based on YOLOv5 \cite{jocher2020yolov5}. The input image is firstly passed through the CSPDarknet53 \cite{wang2020cspnet} backbone which extracts four-scale feature maps \{P3, P4, P5, P6\}. PANet \cite{liu2018path} is used for feature fusion across multi-scales at neck part. Outputs are followed by detection heads. Each final anchor channel prediction is designed as a unified object representation containing box and orientation results.}
	\label{architecture}
\end{figure}

\section{Our Approach}

We adopt the single-stage YOLOv5 \cite{jocher2020yolov5}, which can simultaneously classify objects and regress their locations over a dense grid, as our basic detector. Figure \ref{architecture} illustrates the process of our proposed joint multi-person body detection and orientation estimation (JointBDOE) method. Given an input image, we keep the data augmentation strategy (e.g., Mosaic and MixUp), and utilize the CSPDarknet53 \cite{wang2020cspnet} backbone and PANet \cite{liu2018path} neck for efficient feature extraction and fusion. Then, multi-scale grid heads can predict human instances with various sizes. We have integrated body orientation into the classical object representation. After applying non-maximum suppression (NMS) on these predictions, we finally obtain both location and orientation of all bodies.

\subsection{Unified Embedding}

Generally, we consider the unified embedding to be an extension of the conventional object representation that additionally includes attributes associated with the object. In this way, we can learn multiple related tasks with minimal computational burden for sharing a single network head. 

Here, we suppose the output group of one anchor channel in $i$-th image gird cell with $s \in \{8, 16, 32, 64\}$ times size reduction is $\widehat{\mathcal{H}^s_i}$. In YOLOv5, the anchor channel number $N_a$ of each head is fixed to 3. For one particular anchor channel prediction $\widehat{\mathcal{H}^s_{i,a}}$, its representation can be written as $(\hat{p}, \hat{\bf t}, \hat{\bf c})$. It includes the objectness $\hat{p}$ indicating probability of an object existence, localization offset $\hat{\bf t}=(\hat{x}, \hat{y}, \hat{w}, \hat{h})$ for bounding box, and  classification scores $\hat{\bf c}=(\hat{c}_1, \cdots, \hat{c}_k)$. For our HBOE task, the $\hat{p}$ is typically for the human instance. The object class has only one type ($k=1$). Furthermore, we extend $\widehat{\mathcal{H}^s_{i,a}}$ with one more property $\hat{o}$ denoting body orientation. Now, we obtain a unified embedding $\hat{\bf e}=(\hat{p}, \hat{x}, \hat{y}, \hat{w}, \hat{h}, \hat{c}, \hat{o})$ containing all properties of body (refer Figure \ref{architecture}). Obviously, this unified embedding can be easily evolved to other similar tasks, like encoding Euler angles of eye gaze and head pose.

\subsection{Body Orientation Training}

Human body orientation is defined as one continuous angle $\theta \in [0, 360)$. In our embedding $\hat{\bf e}$, we normalize $o$ into $[0, 1)$ for fitting the $sigmod$ output after each prediction head. During training, we adopt the MSE for body orientation regression. Considering the orientation covers full-range view, we reform a wrapped MSE loss for more reasonable supervision:
\begin{equation}
	\mathcal{L}_{ori}=1/n\sum\nolimits^n_{i=1}\mathsf{min}(\| {\hat{o}_i}-{o}_i \|_2, \|1-|{\hat{o}_i}-o_i|\|_2) ~
\end{equation}
where $\hat{o}_i$ is the estimated result from $i$-th multi-scale head, and $o_i$ is the corresponding ground-truth. The $n$ is 4 here.

\begin{figure}[]
	\centerline{\includegraphics[width=0.7\columnwidth]{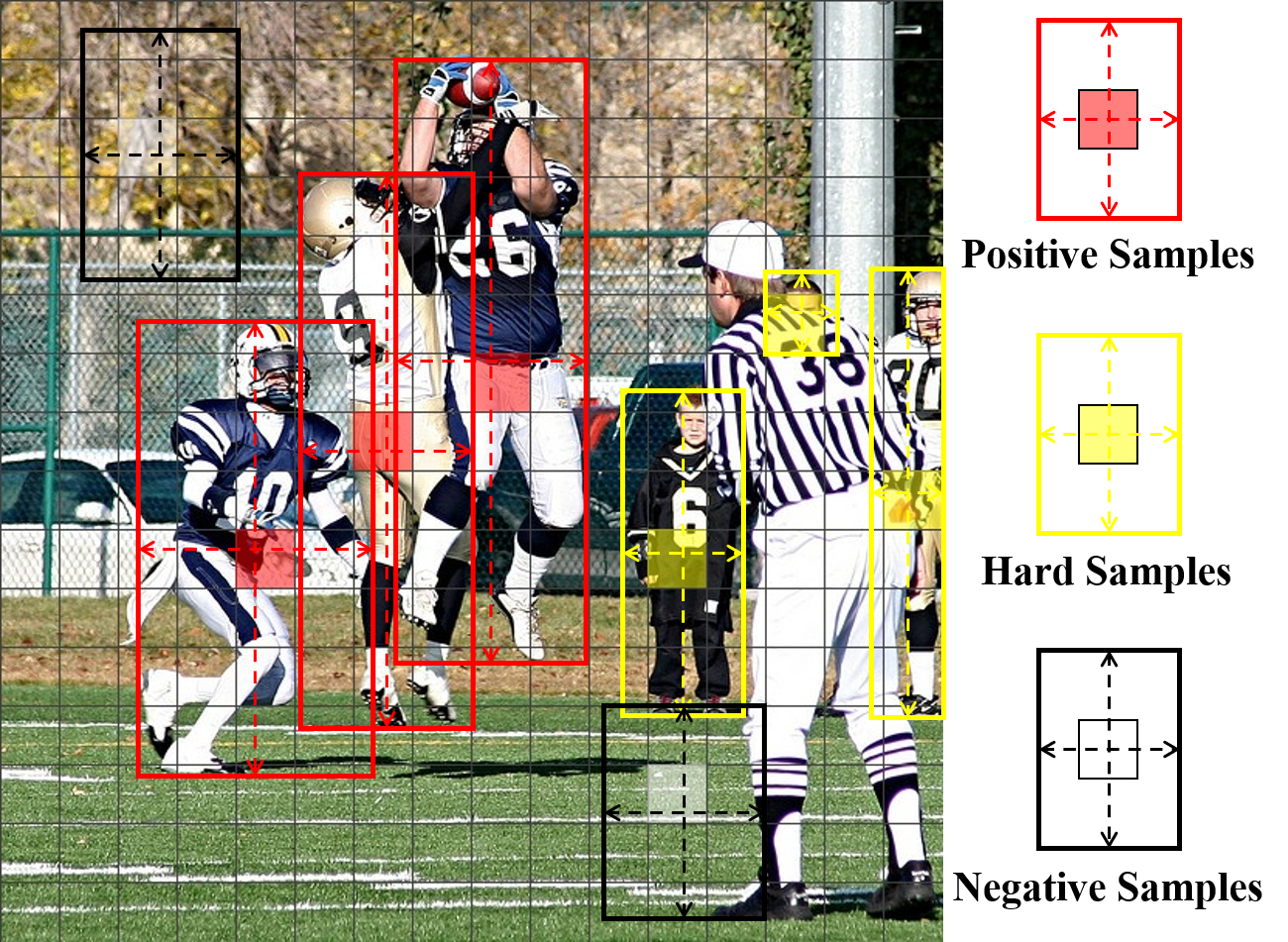}}
	\caption{The illustration of some anchor channel predictions for positive, hard and negative body orientation regression.}
	\label{prediction}
\end{figure}

Although the human body orientation is self-explanatory from the image, we declare that there are two potential troubles: 1) Some severely occluded, highly truncated, or tiny human instances are hard to determine. 2) Dense anchor channel predictions may contain many areas with partial human body or none. Figure \ref{prediction} gives an illustration. These special samples have limited or no effect on the supervised learning of body orientation. Thus, we utilize a tolerance threshold $\tau$ to $\hat{p}$ for filtering out unfavorable $\hat{o}_i$ in each embedding.
\begin{equation}
	\mathcal{L'}_{ori}=\sum\nolimits_s\phi(\hat{p}>\tau)\mathsf{min}(\| \hat{o}-o \|_2, \|1-|\hat{o}-o|\|_2) ~
\end{equation}
However, false negative hard samples should not be discarded and are necessary (refer {\bf Bottom} of Figure \ref{examples}). We obtain a suitable value for the hyper-parameter $\tau$ by ablation studies.

\subsection{Multi-task Loss Optimization}

For body detection training, we follow the original object detection loss design for objectness and localization:
\begin{equation}
	\mathcal{L}_{obj}=1/n\sum\nolimits^n_{i=1}\mathsf{BCE}(\hat{p}, p\cdot\mathsf{CIoU}(\mathbf{\hat{t}_i}, \mathbf{t_i})) ~
\end{equation}
\begin{equation}
	\mathcal{L}_{box}=1/n\sum\nolimits^n_{i=1}[1-\mathsf{CIoU}(\mathbf{\hat{t}_i}, \mathbf{t_i})] ~
\end{equation}
where $\mathsf{BCE}$ is the binary cross-entropy, $\mathsf{CIoU}$ means the complete intersection over union. The body objectness $p=1$ is multiplied by the IoU score to promote concentrated anchor channel predictions. And $p=0$ means no target human body. We do not need the classification loss $\mathcal{L}_{cls}$ for the $\hat{\bf c}$ is meaningless in our HBOE task. The total loss $\mathcal{L}$ is a weighted summation of all three loss components:
\begin{equation}
	\mathcal{L} = \alpha\mathcal{L}_{obj} + \beta\mathcal{L}_{box} + \lambda\mathcal{L'}_{ori} ~
\end{equation}
where we set weights $\alpha=0.7$ and $\beta=0.05$ as in YOLOv5. Then, the optimal value of $\lambda$ for orientation regression loss weight is explored by ablation studies.


\section{Experiments}

\subsection{Implementation Details}

\textbf{Reconstructed MEBOW.}
The MEBOW dataset has labeled 54,007 images, out of which 51,836 images (127,844 human instances) are for training and 2,171 images (5,536 human instances) are for testing. We keep these images and restore challenging human instances that originally given by COCO \cite{lin2014microsoft}, and the corresponding body orientation is given by the method in MEBOW as a weak label. Finally, we have 216,853 and 9,059 instances for training and testing, respectively.

\textbf{Metrics.} Following \cite{wu2020mebow, burgermeister2022pedrecnet}, we report the mean absolute error (MAE) and 
Acc.-$X^{\circ}$ ($X \in \{5, 15, 30\}$) of body orientation estimation results. As for the joint body detection task, we report the AP$^{0.5}$ and Recall results as a reference.

\textbf{Training.} We use YOLOv5s, YOLOv5m and YOLOv5l as our backbones, and follow their basic training settings in \cite{jocher2020yolov5}. On our reconstructed MEBOW dataset, the max epoch is set to 500. The input images are resized and padded to $1024 \times 1024$, keeping the original aspect ratio. Training parameters $\tau$ and $\lambda$ are manually tuned by experiments.

\setlength{\tabcolsep}{3.2pt}
\begin{table}[!t]\scriptsize 
\begin{center}
\caption{Influence exploration of two hyperparameters: the tolerance threshold $\tau$ and the loss weight $\lambda$ for $\mathcal{L'}_{ori}$.}
\label{ablation}
\begin{tabular}{c|c|c|ccc|cc}
\Xhline{1.5pt}
$\tau$ & $\lambda$ & {\bf MAE}$\downarrow$ & {\bf Acc.-$5^{\circ}$}$\uparrow$ & {\bf Acc.-$15^{\circ}$}$\uparrow$ & {\bf Acc.-$30^{\circ}$}$\uparrow$ & {\bf AP}$^{0.5}$$\uparrow$ & {\bf AP}$^{.5:.95}$$\uparrow$ \\
\Xhline{1pt}
0.0 & 0.1 & 14.031 & 42.96 & 78.70 & 90.35 & 82.4 & 58.4 \\
0.1 & 0.1 & 13.848 & 43.60 & 79.29 & 90.83 & 82.9 & 59.0 \\
\rowcolor{gray!20}
0.2 & 0.1 & {\bf 13.758} & 43.02 & 79.02 & 90.67 & 82.8 & 58.7 \\
0.3 & 0.1 & 14.493 & 41.98 & 77.91 & 89.68 & 82.3 & 58.1 \\
0.4 & 0.1 & 15.122 & 40.66 & 76.84 & 89.69 & 81.7 & 57.4 \\
0.5 & 0.1 & 15.247 & 41.26 & 77.02 & 89.03 & 81.6 & 57.2 \\
\Xhline{1pt}
0.2 & 0.02 & 13.738 & 41.44 & 79.18 & 90.88 & 84.8 & 62.0  \\
\rowcolor{gray!20}
0.2 & 0.05 & {\bf 13.427} & 42.88 & 79.64 & 91.21 & 84.1 & 61.1  \\
0.2 & 0.10 & 13.758 & 43.02 & 79.02 & 90.67 & 82.8 & 58.7 \\
0.2 & 0.15 & 14.715 & 41.08 & 77.21 & 89.96 & 80.5 & 55.7  \\
\Xhline{1.5pt}
\end{tabular}
\end{center}
\vspace{-15pt}
\end{table}

\subsection{Evaluation on Datasets}

\textbf{Ablation Studies.} For simplicity, we used YOLOv5s to train 300 epochs, and the last model to test for finding optimal parameters. In Table \ref{ablation}, we temporarily set $\lambda$ to 0.1 and selected threshold $\tau$ from 0.0 to 0.5 with a step 0.1. The lowest MAE is obtained when $\tau$ is 0.2, which indicates an appropriate filtering of hard samples is vital. Then, we fixed $\tau$ and chose the loss weight $\lambda$ for $\mathcal{L'}_{ori}$ from $\{0.02, 0.05, 0.10, 0.15\}$. When $\lambda$ is 0.05, we got a better trade-off between body detection and orientation estimation.

\textbf{Comparison.} Finally, we quantitatively and qualitatively demonstrate the impressive results achieved by our proposed method. As shown in Table \ref{compare}, our method trained on reconstructed MEBOW (no mark$^{\dagger}$) achieved high AP$^{0.5}$ and Recall of body detection and reasonable body orientation accuracy. For a fair comparison, we also evaluated our models on the original MEBOW discarding many challenging instances (with mark$^{\dagger}$). The model using YOLOv5l$^{\dagger}$ obtained approximative MAE and accuracy results with MEBOW \cite{wu2020mebow} and PedRecNet \cite{burgermeister2022pedrecnet} which are elaborately designed and firmly dedicated to the single HBOE task. Thus, considering that our method is weakly supervised learning based and trained on the whole original image toward multi-persons, such results with relative disparity are comparable.

\begin{figure}[]
\begin{minipage}[b]{.495\linewidth}
  \centering
  \centerline{\includegraphics[width=\columnwidth]{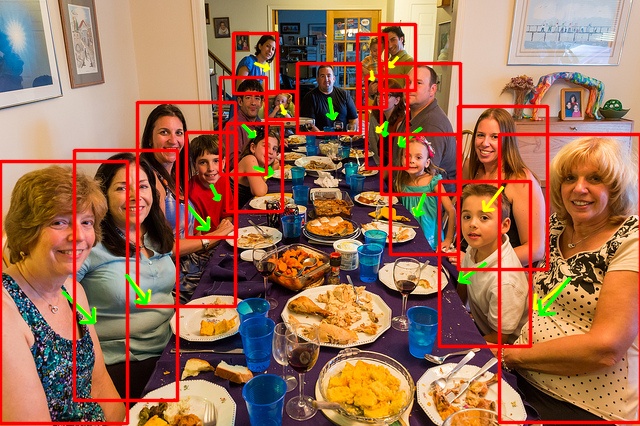}}
\end{minipage}
\hfill
\begin{minipage}[b]{.495\linewidth}
  \centering
  \centerline{\includegraphics[width=\columnwidth]{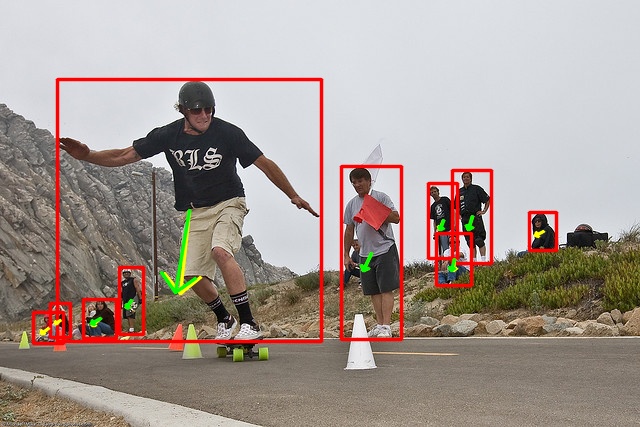}}
\end{minipage}
\begin{minipage}[b]{.495\linewidth}
  \centering
  \centerline{\includegraphics[width=\columnwidth]{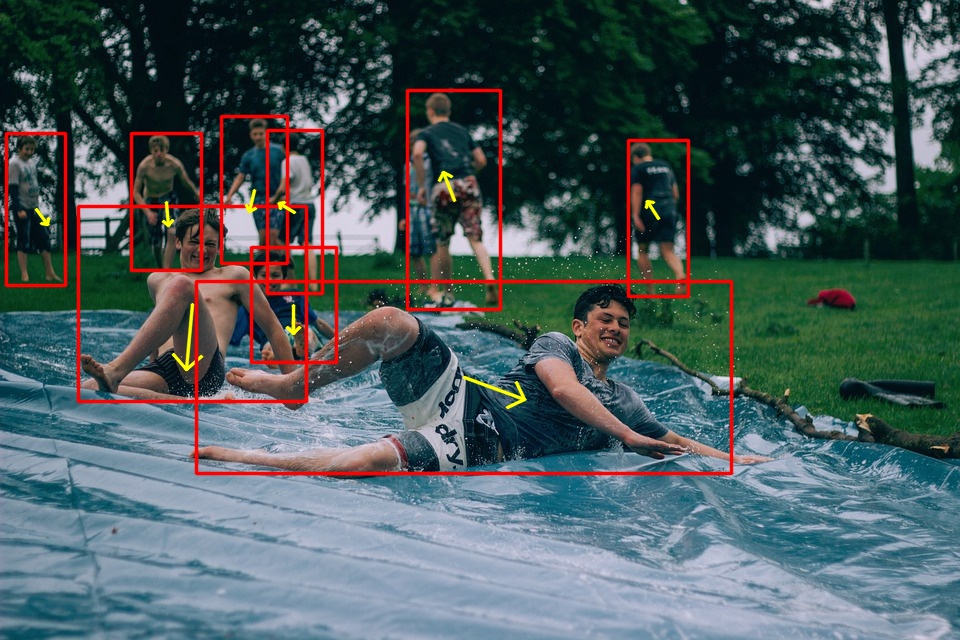}}
\end{minipage}
\hfill
\begin{minipage}[b]{.495\linewidth}
  \centering
  \centerline{\includegraphics[width=\columnwidth]{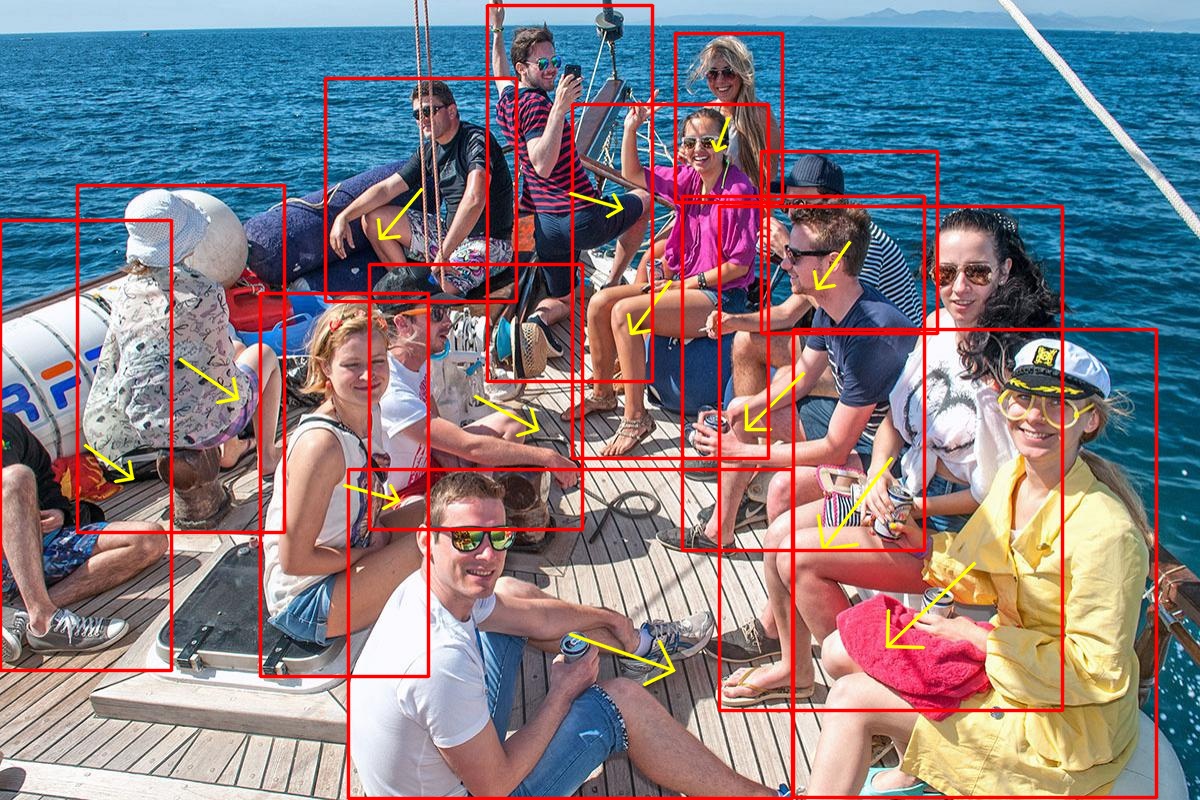}}
\end{minipage}
\caption{Qualitative examples. {\bf Top:} Images in MEBOW \cite{lin2014microsoft} val-set with HBOE labels (some instances are unlabeled). {\bf Bottom:} Images in CrowdHuman \cite{shao2018crowdhuman} without body orientation labels. Green arrow: labels; Yellow arrow: predictions.}
\label{qualitative}
\vspace{-5pt}
\end{figure}

Some qualitative examples of our model using YOLOv5l are presented in Figure \ref{qualitative}. Predictions on labeled MEBOW are closed to their real orientations. A few unlabeled tiny or occluded bodies can also be detected and reasonably estimated. Our method is also quite robust to blurry and crowded scenes from dataset CrowdHuman. These are largely attributable to the joint learning advantages of the body detection task.

\setlength{\tabcolsep}{1pt}
\begin{table}[!t]\scriptsize 
\begin{center}
\caption{Body orientation performance comparison of our method with SOTA methods in MEBOW dataset. The ''---'' means not reported. Mark$^{\dagger}$ means that we evaluate our models on original val-set of MEBOW for a fair consideration.}
\label{compare}
\begin{tabular}{l|l|c|ccccc|c|c}
\Xhline{1.5pt}
& \multirow{2}{*}{\bf Methods} & \multirow{2}{*}{\bf MAE$\downarrow$} & & & {\bf Accuracy}$\uparrow$ & & & \multirow{2}{*}{\bf AP$^{0.5}\uparrow$} & \multirow{2}{*}{\bf Recall$\uparrow$} \\
& & & $5^{\circ}$ & $15^{\circ}$ & $22.5^{\circ}$ & $30^{\circ}$ & $45^{\circ}$ &  & \\
\Xhline{1pt}
\multirow{2}{*}{\makecell{{\bf Single-}\\{\bf person}}} & MEBOW \cite{wu2020mebow} & {\bf 8.393} & 68.6 & 90.7 & 93.9 & 96.9 & 98.2 & --- & 100.0 \\
& PedRecNet \cite{burgermeister2022pedrecnet} & 9.700 & --- & --- & 92.3 & --- & 97.0 & --- &100.0 \\
\hline
\multirow{6}{*}{\makecell{{\bf Multi-}\\{\bf person}}} & Ours-YOLOv5s & 20.670 & 37.0 & 69.7 & 79.0 & 83.5 & 87.9 & 84.4 & 97.37 \\ 
& Ours-YOLOv5m & 18.763 & 39.8 & 73.5 & 81.4 & 85.5 & 89.5 & 85.8  & 97.63 \\
& Ours-YOLOv5l & 17.848 & 40.8 & 74.4 & 82.3 & 86.5 & 90.4 & 86.1 & 97.57 \\
\cline{2-10}
& Ours-YOLOv5s$^{\dagger}$ & 13.130 & 44.1 & 79.9 & 88.1 & 91.8 & 94.2 & --- & 99.55 \\
& Ours-YOLOv5m$^{\dagger}$ & 11.907 & 46.4 & 83.0 & 89.7 & 92.9 & 95.4 & --- & 99.49 \\
& Ours-YOLOv5l$^{\dagger}$ & {\bf 11.207} & 47.5 & 83.4 & 90.5 & 93.8 & 96.0 & --- & 99.46 \\ 
\Xhline{1.5pt}
\end{tabular}
\end{center}
\vspace{-5pt}
\end{table}

\section{Conclusions}

In this paper, we propose a novel single-stage joint human body detection and orientation estimation method for multi-person scenes. In order to share the features extracted by body detection, we subtly expand the object representation and design a unified embedding including the body orientation attribute. Observing that not all anchor channel predictions contribute orientation regression consistently, we filter out undesired samples in its loss. Finally, our method performs well on the reconstructed MEBOW dataset for the multi-person HBOE task. Moreover, the near real-time efficiency of our model is not affected by the number of people in the image.


\vfill\pagebreak

\bibliographystyle{IEEEbib}
\bibliography{refs}

\begin{thebibliography}{10}

\bibitem{rehder2014head}
Eike Rehder, Horst Kloeden, and Christoph Stiller,
\newblock ``Head detection and orientation estimation for pedestrian safety,''
\newblock in {\em ITSC}. IEEE, 2014, pp. 2292--2297.

\bibitem{yu2019continuous}
Dameng Yu, Hui Xiong, Qing Xu, Jianqiang Wang, and Keqiang Li,
\newblock ``Continuous pedestrian orientation estimation using human
  keypoints,''
\newblock in {\em ISCAS}, 2019, pp. 1--5.

\bibitem{araya2021automatic}
Roberto Araya and Jorge Sossa-Rivera,
\newblock ``Automatic detection of gaze and body orientation in elementary
  school classrooms,''
\newblock {\em Frontiers in Robotics and AI}, p. 277, 2021.

\bibitem{ricci2015uncovering}
Elisa Ricci, Jagannadan Varadarajan, Ramanathan Subramanian, Samuel Rota~Bulo,
  Narendra Ahuja, and Oswald Lanz,
\newblock ``Uncovering interactions and interactors: Joint estimation of head,
  body orientation and f-formations from surveillance videos,''
\newblock in {\em ICCV}, 2015, pp. 4660--4668.

\bibitem{raza2018appearance}
Mudassar Raza, Zonghai Chen, Saeed-Ur Rehman, Peng Wang, and Peng Bao,
\newblock ``Appearance based pedestrians’ head pose and body orientation
  estimation using deep learning,''
\newblock {\em Neurocomputing}, vol. 272, 2018.

\bibitem{wu2020mebow}
Chenyan Wu, Yukun Chen, Jiajia Luo, Che-Chun Su, Anuja Dawane, Bikramjot
  Hanzra, Zhuo Deng, Bilan Liu, James~Z Wang, and Cheng-hao Kuo,
\newblock ``Mebow: Monocular estimation of body orientation in the wild,''
\newblock in {\em CVPR}, 2020, pp. 3451--3461.

\bibitem{andriluka2010monocular}
Mykhaylo Andriluka, Stefan Roth, and Bernt Schiele,
\newblock ``Monocular 3d pose estimation and tracking by detection,''
\newblock in {\em CVPR}. IEEE, 2010, pp. 623--630.

\bibitem{baltieri2012people}
Davide Baltieri, Roberto Vezzani, and Rita Cucchiara,
\newblock ``People orientation recognition by mixtures of wrapped distributions
  on random trees,''
\newblock in {\em ECCV}. Springer, 2012, pp. 270--283.

\bibitem{choi2016human}
Jinyoung Choi, Beom-Jin Lee, and Byoung-Tak Zhang,
\newblock ``Human body orientation estimation using convolutional neural
  network,''
\newblock {\em arXiv preprint arXiv:1609.01984}, 2016.

\bibitem{hara2017growing}
Kota Hara and Rama Chellappa,
\newblock ``Growing regression tree forests by classification for continuous
  object pose estimation,''
\newblock {\em IJCV}, vol. 122, no. 2, pp. 292--312, 2017.

\bibitem{liu2017weighted}
Peiye Liu, Wu~Liu, and Huadong Ma,
\newblock ``Weighted sequence loss based spatial-temporal deep learning
  framework for human body orientation estimation,''
\newblock in {\em ICME}. IEEE, 2017, pp. 97--102.

\bibitem{liu2013accurate}
Wu~Liu, Yongdong Zhang, Sheng Tang, Jinhui Tang, Richang Hong, and Jintao Li,
\newblock ``Accurate estimation of human body orientation from rgb-d sensors,''
\newblock {\em IEEE Transactions on cybernetics}, vol. 43, no. 5, pp.
  1442--1452, 2013.

\bibitem{lin2014microsoft}
Tsung-Yi Lin, Michael Maire, Serge Belongie, James Hays, Pietro Perona, Deva
  Ramanan, Piotr Doll{\'a}r, and C~Lawrence Zitnick,
\newblock ``Microsoft coco: Common objects in context,''
\newblock in {\em ECCV}. Springer, 2014, pp. 740--755.

\bibitem{ren2015faster}
Shaoqing Ren, Kaiming He, Ross Girshick, and Jian Sun,
\newblock ``Faster r-cnn: Towards real-time object detection with region
  proposal networks,''
\newblock {\em NIPS}, vol. 28, 2015.

\bibitem{tian2019fcos}
Zhi Tian, Chunhua Shen, Hao Chen, and Tong He,
\newblock ``Fcos: Fully convolutional one-stage object detection,''
\newblock in {\em ICCV}, 2019, pp. 9627--9636.

\bibitem{cao2017realtime}
Zhe Cao, Tomas Simon, Shih-En Wei, and Yaser Sheikh,
\newblock ``Realtime multi-person 2d pose estimation using part affinity
  fields,''
\newblock in {\em CVPR}, 2017, pp. 7291--7299.

\bibitem{jocher2020yolov5}
Glenn Jocher, K~Nishimura, T~Mineeva, and R~Vilari{\~n}o,
\newblock ``Yolov5,''
\newblock {\em Code repository https://github.com/ultralytics/yolov5}, 2020.

\bibitem{zhang2016joint}
Kaipeng Zhang, Zhanpeng Zhang, Zhifeng Li, and Yu~Qiao,
\newblock ``Joint face detection and alignment using multitask cascaded
  convolutional networks,''
\newblock {\em SPL}, vol. 23, no. 10, pp. 1499--1503, 2016.

\bibitem{wu2021synergy}
Cho-Ying Wu, Qiangeng Xu, and Ulrich Neumann,
\newblock ``Synergy between 3dmm and 3d landmarks for accurate 3d facial
  geometry,''
\newblock in {\em 3DV}. IEEE, 2021, pp. 453--463.

\bibitem{burgermeister2022pedrecnet}
Dennis Burgermeister and Crist{\'o}bal Curio,
\newblock ``Pedrecnet: Multi-task deep neural network for full 3d human pose
  and orientation estimation,''
\newblock {\em IV}, 2022.

\bibitem{nonaka2022dynamic}
Soma Nonaka, Shohei Nobuhara, and Ko~Nishino,
\newblock ``Dynamic 3d gaze from afar: Deep gaze estimation from temporal
  eye-head-body coordination,''
\newblock in {\em CVPR}, 2022, pp. 2192--2201.

\bibitem{zhang2017citypersons}
Shanshan Zhang, Rodrigo Benenson, and Bernt Schiele,
\newblock ``Citypersons: A diverse dataset for pedestrian detection,''
\newblock in {\em CVPR}, 2017, pp. 3213--3221.

\bibitem{shao2018crowdhuman}
Shuai Shao, Zijian Zhao, Boxun Li, Tete Xiao, Gang Yu, Xiangyu Zhang, and Jian
  Sun,
\newblock ``Crowdhuman: A benchmark for detecting human in a crowd,''
\newblock {\em arXiv preprint arXiv:1805.00123}, 2018.

\bibitem{wang2020cspnet}
Chien-Yao Wang, Hong-Yuan~Mark Liao, Yueh-Hua Wu, Ping-Yang Chen, Jun-Wei
  Hsieh, and I-Hau Yeh,
\newblock ``Cspnet: A new backbone that can enhance learning capability of
  cnn,''
\newblock in {\em CVPRW}, 2020, pp. 390--391.

\bibitem{liu2018path}
Shu Liu, Lu~Qi, Haifang Qin, Jianping Shi, and Jiaya Jia,
\newblock ``Path aggregation network for instance segmentation,''
\newblock in {\em CVPR}, 2018, pp. 8759--8768.

\end{thebibliography}

\end{document}